%% file: main.tex
\documentclass[10pt,twocolumn,letterpaper]{article}

\usepackage[pagenumbers]{cvpr} % To force page numbers, e.g. for an arXiv version


\definecolor{cvprblue}{rgb}{0.21,0.49,0.74}
\usepackage[pagebackref,breaklinks,colorlinks,allcolors=cvprblue]{hyperref}

\usepackage[accsupp]{axessibility}  % Improves PDF readability for those with disabilities.

\def\paperID{36} % *** Enter the Paper ID here
\def\confName{CVPR}
\def\confYear{2025}

\title{CytoFM: The first cytology foundation model}

\author{Vedrana Ivezić\\
University of California, Los Angeles, USA\\
{\tt\small vivezic@g.ucla.edu}
\and
Ashwath Radhachandran\\
University of California, Los Angeles, USA\\
{\tt\small ashwathradha123@g.ucla.edu}
\and
Ekaterina Redekop\\
University of California, Los Angeles, USA\\
{\tt\small eredekop@g.ucla.edu}
\and
Shreeram Athreya\\
University of California, Los Angeles, USA\\
{\tt\small shreeram@ucla.edu}
\and
Dongwoo Lee\\
University of California, Los Angeles, USA\\
{\tt\small rlee8517@g.ucla.edu}
\and
Vivek Sant\\
UT Southwestern Medical Center, Dallas, TX\\
{\tt\small vivek.sant@utsouthwestern.edu}
\and
Corey Arnold\\
University of California, Los Angeles, USA\\
{\tt\small cwarnold@ucla.edu}
\and
William Speier\\
University of California, Los Angeles, USA\\
{\tt\small Speier@ucla.edu}
}

\begin{document}
\maketitle
\input{sec/0_abstract}
\input{sec/1_intro}
\input{sec/2_methods}

\input{sec/3_experiments}
\input{sec/4_conclusions}
{
    \small
    \bibliographystyle{ieeenat_fullname}
    \bibliography{main}
}


\end{document}

%% file: sec/0_abstract.tex
\begin{abstract}
Cytology is essential for cancer diagnostics and screening due to its minimally invasive nature. However, the development of robust deep learning models for digital cytology is challenging due to the heterogeneity in staining and preparation methods of samples, differences across organs, and the limited availability of large, diverse, annotated datasets. Developing a task-specific model for every cytology application is impractical and non-cytology-specific foundation models struggle to generalize to tasks in this domain where the emphasis is on cell morphology. To address these challenges, we introduce CytoFM, the first cytology self-supervised foundation model. Using iBOT, a self-supervised Vision Transformer (ViT) training framework incorporating masked image modeling and self-distillation, we pretrain CytoFM on a diverse collection of cytology datasets to learn robust, transferable representations. We evaluate CytoFM on multiple downstream cytology tasks, including breast cancer classification and cell type identification, using an attention-based multiple instance learning framework. Our results demonstrate that CytoFM performs better on two out of three downstream tasks than existing foundation models pretrained on histopathology (UNI) or natural images (iBOT-Imagenet). Visualizations of learned representations demonstrate our model is able to attend to cytologically relevant features. Despite a small pre-training dataset, CytoFM's promising results highlight the ability of task-agnostic pre-training approaches to learn robust and generalizable features from cytology data.
\end{abstract}

%% file: sec/1_intro.tex
\section{Introduction}
\label{sec:intro}
Cytology, also known as cytopathology, is a specialty of pathology which examines cells from bodily tissues or fluids using a microscope for the purpose of screening and diagnosing cancer \cite{al-abbadi_basics_2011}. Collecting a cytology biopsy sample compared to a surgical biopsy is safer, faster, and more cost-effective \cite{al-abbadi_basics_2011}. Cytology tests are used as a screening tool in cancer diagnostics and for conducting follow-ups on suspicious new lesions or fluids from patients with a previous cancer diagnosis \cite{al-abbadi_basics_2011}. The standard preparation of a cytology sample begins with smearing the sample onto a slide, staining the specimen for cell visualization, and finally, evaluation by a cytopathologist. Unlike histopathology, which focuses on tissue and cell architecture, cytology examines individual cells for their morphology, cohesiveness, and other cellular properties \cite{al-abbadi_basics_2011,giarnieri_towards_2023}. As a result, a cytopathologist must evaluate each cell under a microscope to identify any diagnostically important alterations which often are present in very few of the examined cells \cite{layfield_what_1997}. Given the sparsity of diagnostic cells in the large slides, evaluation of cytology slides is a labor-intensive and expensive task which can delay diagnosis reporting \cite{renshaw_time_2016}. Recent work has aimed to automate the diagnosis process through deep learning using digitized cytology slides, referred to as whole slide images (WSIs) \cite{zhao_less_2024,dov_weakly_2021,landau_artificial_2019, marletta_whole-slide_2021}. 

Due to their gigapixel size, WSIs are unable to fit at full resolution into deep learning models. Instead, patch-based methods are commonly used to divide the WSIs into smaller images which are more feasible for model training \cite{zhang_deeppap_2017,zhao_less_2024,dov_weakly_2021}. Because patch-level labels are often challenging and time-consuming to obtain, multiple instance learning (MIL) is used to aggregate patch level information and is trained on slide-level or patient-level labels as weak supervision \cite{dov_weakly_2021}. 

%Learning robust representations of the patches is key to developing well-performing diagnostic models. 

Previous techniques to develop models to learn robust representations of cytology images involve training and evaluating convolutional neural networks on a single dataset. Due to the variations across cytology data, a model trained on a single dataset will struggle to generalize to external datasets \cite{welch_bmt_2024,plissiti_sipakmed_2018}. While techniques like stain normalization can be used to address these variations, these methods are not specific to cytology stains and are often not readily available. Rather than relying on normalization techniques, Huang et al. used foundation models which are less susceptible to noise for cytology feature extraction \cite{huang_efficient_2024}. They proposed to use a frozen feature extractor from foundation models such as BiomedCLIP with a trainable adapter added to the extractor output. The adapter was then trained with parameter-efficient fine-tuning using contrastive learning to increase the specificity of the foundation model features to the task, such as cervical cancer classification. However, because this model was trained to adapt to individual datasets, it must be fully retrained to apply to other datasets and tasks. Using foundation models which are not cytology-specific as the feature extractor and performing task-specific fine-tuning with labeled datasets have also shown promising results \cite{dausort_exploring_2024}. However, Dausort et al. found that histopathology-trained foundation models struggled to generalize to cytology data, likely due to the focus on tissue structure in histology which is not present in cytology \cite{dausort_exploring_2024}. Due to these foundation models not being trained on cytology datasets, they were not able to fully represent and utilize the information present in cytology images.

Unlike previous approaches which rely on explicit annotations, self-supervised learning learns the inherent structure in unlabeled data through pretext tasks, resulting in meaningful representations generalizable to unseen domains. In the field of histopathology where similar WSI related challenges arise, there has been a surge in self-supervised foundation models trained on large and diverse unlabeled datasets \cite{chen_scaling_2022,vorontsov_foundation_2024,chen_towards_2024}. Recent studies have applied self-distillation and masked image modeling to improve the representation learning of Vision Transformers (ViTs) in histology WSIs. These models have proven to be strong feature encoders and robust to differences in staining, preparation style, and other institution-specific differences \cite{vorontsov_foundation_2024,chen_towards_2024}. 

Self-supervised foundation models have not been well explored in cytology primarily due to the limited availability of public cytology datasets. Dausort et al. proposed using four labeled public cytology datasets to fine tune existing foundation models pretrained on medical data. This approach does not fully utilize the publicly available cytology datasets because of the reliance on labels and only uses a single dataset for fine-tuning. With the recent release of new publicly available cytology datasets and the use of a private thyroid cytology dataset, we propose using self-supervised learning to develop a strong and generalizable cytology feature extractor. Self-supervised learning allows the use of both unlabeled and labeled datasets and can learn robust representations from diverse and noisy data. Such an approach can leverage the available cytology datasets to develop a model capable of learning robust, cytology-specific representations.

%Therefore, a cytology-specific foundation model trained on unlabeled cytology datasets using self-supervised learning serves as a generalizable feature extractor for cytology by learning broad, non-task-specific features. Such a model improves downstream performance while also reducing the necessity for extensive, labor-intensive training data.

\begin{figure*}[t]
  \centering
   %\includegraphics[width=0.8\linewidth]{cytofm.jpg}
   \includegraphics[width=0.8\textwidth]{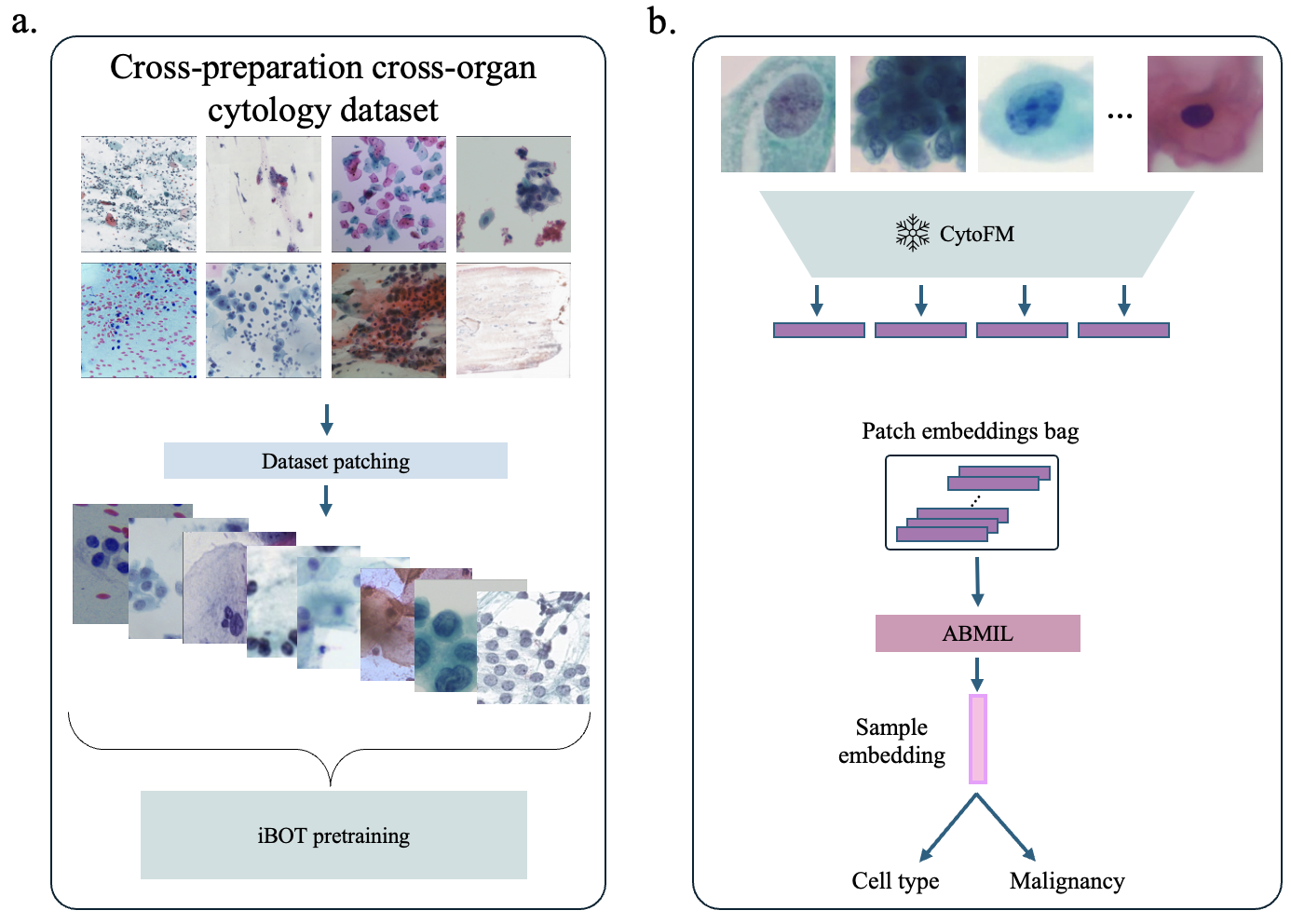}

   \caption{\textbf{CytoFM}: the first cytology specific foundation model. Developing this model requires the curation of an unlabeled cytology specific dataset which is used to pre-train a ViT, using the iBOT framework (a). The trained ViT, CytoFM, is then used to extract features for cytology image patches; the patch features for an image are aggregated into a bag and an ABMIL framework creates a single embedding for an image to be used in downstream tasks (b).}
   \label{fig:cytofm}
\end{figure*}

In this work, we introduce CytoFM (figure \ref{fig:cytofm}), the first cytology-specific foundation model. CytoFM is trained on a diverse set of public and private datasets using self-supervised learning. We pretrain a ViT-Base model by leveraging iBOT, a self-supervised framework for pre-training vision transformers (ViT) yielding representations which do not need further fine-tuning \cite{zhou_ibot_2022}. iBOT uses both masked image modeling and self-distillation learning to learn high- and low-level features represented in a single image embedding. iBOT's rich representation is necessary for cytology where both low-level information, such as mitotic activity, as well as high-level information, like cellularity, are important in diagnostics. We evaluate our model on downstream tasks including breast cancer diagnosis and cell type classification using an attention multiple instance learning (ABMIL) framework. Performance results are compared against both an iBOT model trained on ImageNet (iBOT-Imagenet) \cite{zhou_ibot_2022} and a pathology-specific foundation model, UNI \cite{chen_towards_2024}. We summarize our contributions as follows:
\begin{itemize}
  \item CytoFM is the first foundation model specifically designed for cytology 
  \item Our model learns robust representations of cytology images,  capturing important features including cell morphology and mitotic activity 
  \item Despite the limited dataset size, we demonstrate that self-supervised learning on a collection of diverse cytology datasets performs better on two out of three downstream tasks compared to non cytology-specific models in feature extraction and generalization to unseen datasets.
  %\item CytoFM serves as an initial proof of concept for cytology specific foundation models. The success of CytoFM given the dataset limitations underscores the potential for future iterations trained on larger and more diverse datasets to establish a new standard for cytology deep learning models. 
\end{itemize}

%-------------------------------------------------------------------------

%-------------------------------------------------------------------------

%-------------------------------------------------------------------------

%% file: sec/2_methods.tex
\section{Methods}
\label{sec:Methods}

%-------------------------------------------------------------------------
\subsection{Datasets}
\label{subsec:Datasets}
Foundation models require a large amount of diverse data to learn meaningful and generalizable representations. We combine seven publicly available datasets and one private dataset, creating a multi-institutional dataset comprising 1,393,282 patches which span three types of organs including breast, cervix, and thyroid for pretraining our foundation model (table \ref{table_datasets}). 

%These crops are chosen by cytologists as containing cells and areas of cytological importance and are still relatively large images requiring patching to be performed.

The breast cytology dataset we include is the \textbf{Saikia et al. fine needle aspiration} (FNAC2019) dataset \cite{saikia_comparative_2019}. Papanicolau (Pap) stained breast cytology slides were digitized using an ICC50 HD microscope and certified cytopathologists selected and labeled 212 images as benign or malignant.

We include five different publicly available cervical cytology datasets in the pretraining phase. 

\begin{itemize}
    \item The \textbf{Mendeley LBC Cervical Cancer} dataset (MLBC) \cite{hussain_liquid_2020} is a liquid-based cytology cervical cancer dataset of 963 Pap smear images at 40x magnification collected from 460 patients using a Leica ICC50 HD microscope. Image level labels for the subcategories of cervical cancer lesions as defined by The Bethesda System for Reporting Cervical Cytology are included.
    \item The \textbf{SiPaKMeD} dataset contains \cite{plissiti_sipakmed_2018} 4049 manually cropped images from cervical Pap smear slides imaged using an Infinity 1 Lumenera CCD camera adapted to an OLYMPUS BX53F optical microscope. Each image is labeled by expert cytopathologists as one of five cell types, which span normal, abnormal, and benign categories.
    \item The \textbf{Brown Multicellular ThinPrep Database} (BMT) \cite{welch_bmt_2024} consists of 600 clinician selected images taken from 180 cervical ThinPrep\textsuperscript{\textregistered} Pap smear slides from 180 patients. Images were obtained using an Olympus BX43 microscope and an Excelis HD color microscopy camera, and were labeled as one of three diagnostic cell types by three board-certified pathologists.
    \item The \textbf{Annotated Pap smear images for Cell Segmentation 2023} (APACS23) \cite{harangi_pixel-wise_2024} is a cervical Pap smear image dataset composed of 3565 images curated from five cytology slides from five patients with cell segmentations provided. The slides were digitized using a 3DHistec Pannoramic 1000 digital slide scanner and an Adimec Q-12A-180Fc brightfield camera.
    \item The \textbf{Cervical Cell Edge Detection} (CCEDD) \cite{liu_local_2022} dataset contains 686 images acquired by a Nikon ELIPSE Ci slide scanner, a SmartV350D lens, and a 3-megapixel digital camera, with expert cytologist cell annotations. 
    \item The \textbf{Bialystok} cervical cytology dataset \cite{pater_conventional_2023} contains 162 images from 20 Pap stained slides. Cell type annotations by cytotechnologists were performed for each image, classifying cells into Bethesda defined categories. 
\end{itemize}

\begin{table*}[]
\centering
\begin{tabular}{c|c|c|c|c}
\textbf{Dataset name} & \textbf{Label}                     & \textbf{Organ}   & \textbf{Pre-training} & \textbf{Evaluation} \\ \hline \hline
FNAC2019     & binary                    & Breast  & x               & x          \\ \hline
MLBC         & multiclass                & Cervix  & x               & x          \\ \hline
SiPaKMeD     & multiclass                & Cervix  & x               &            \\ \hline
BMT          & multiclass                & Cervix  & x               &            \\ \hline
APACS23      & segmentations             & Cervix  & x               &            \\ \hline
CCEDD        & segmentations             & Cervix  & x               &            \\ \hline
Bialystok    & multiclass, segmentations & Cervix  & x               &            \\ \hline
HiCervix     & multiclass                & Cervix  &                 & x          \\ \hline
ThyUCLA      & binary                    & Thyroid & x               &            \\ \hline
\end{tabular}
\caption{\textbf{Description of datasets} used in the  development and evaluation of CytoFM. Datasets are marked if they were used in pre-training the model and if they are also used in evaluation. If a dataset is used in both pre-training and evaluation, it is split into two sets.
}
\label{table_datasets}
\end{table*}

We also include a private dataset of \textbf{thyroid cytology WSIs} from our institution to increase the size and diversity of the dataset. Our dataset contains 496 Pap stained slides from 160 patients digitized with a Leica Aperio GT450 scanner at 40x magnification. The cohort consists of patients with an indeterminate cytology as defined by the Bethesda System for Reporting Thyroid Cytopathology, followed by a suspicious molecular test and surgical resection which provides a final surgical pathology diagnosis at the patient level.

To evaluate the performance and generalizability of CytoFM, we include a large public dataset for downstream evaluation which is completely unseen during the CytoFM training process. The \textbf{HiCervix} dataset \cite{cai_hicervix_2024} contains 4,496 Pap stained WSIs which were digitized using either the ZEISS Primostar 3, Olympus microscope BX43, or Sunnyoptic RX50 microscope along with a CCD camera. Images of 40,229 cervical cells were extracted from the WSIs and categorized into a three-level hierarchical tree with 29 cell types included.

%-------------------------------------------------------------------------

\begin{figure}[t]
  \centering
   \includegraphics[width=0.8\linewidth]{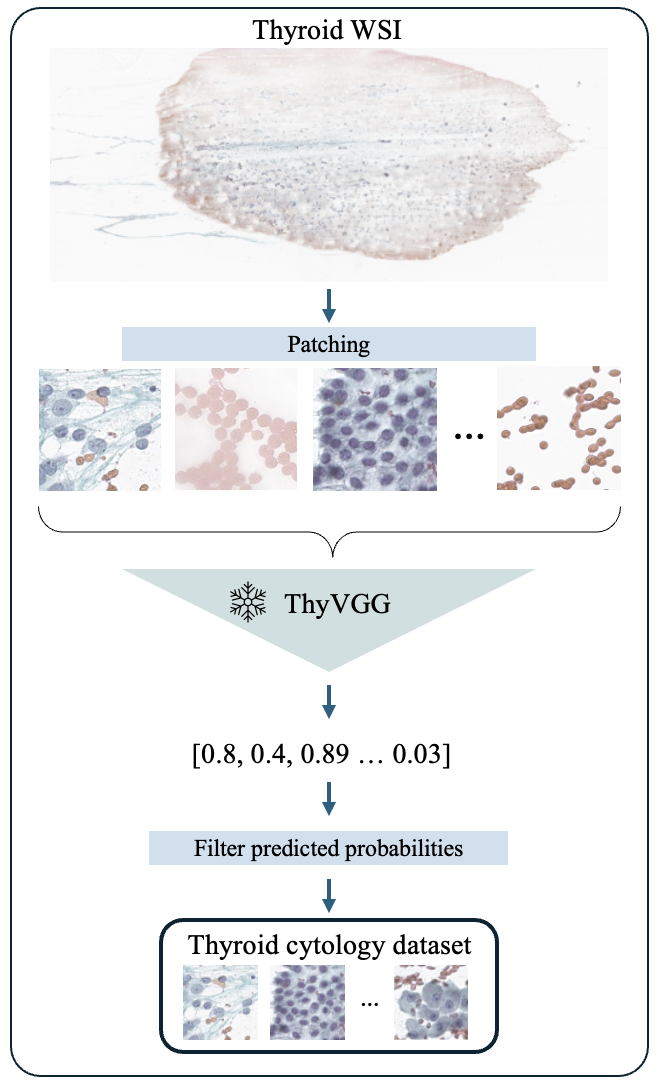}

   \caption{\textbf{Filtering of patches from a WSI} for our private thyroid dataset. A WSI is patched and ThyVGG is used to predict the probability that the patch contains relevant information. The top 1500 patches with the highest probability from a slide are used in the thyroid cytology dataset.}
   \label{fig:thyroid_filtering}
\end{figure}

%-------------------------------------------------------------------------
\subsection{Preprocessing}
\label{subsec:preprocess}

To ensure consistency in size and magnification, images from each dataset are split into non-overlapping 256x256 pixel patches at 40x magnification. Patches at the edges of images are zero padded or dropped if more than half of the patch requires padding. In the case where a dataset does not provide the magnification level, we calculate it based on the average pixel size of a nucleus compared to the average micrometer size of a nucleus. Using the calculated magnification we rescale the image to 40x magnification. While histopathology foundation models generally patch WSIs at 20x magnification \cite{chen_towards_2024}, we choose a higher magnification because cytology focuses on cellular features and small clusters of cells rather than broader cell and tissue structure. 

Because cytology slides contain sparse cells, removing slide preparation debris and background areas is essential for creating an informative dataset that allows the model to learn meaningful representations. In the public datasets we used, this was achieved by previous preprocessing of WSIs into smaller images of cytologically relevant cells by cytologists. However, in our private dataset, this was not possible. To achieve comparable preprocessing, we employ our ThyVGG model (figure \ref{fig:thyroid_filtering}) to predict the probability of a patch containing diagnostically relevant information. ThyVGG is a VGG-16 model initialized with ImageNet weights and fine-tuned on a set of 5000 patches annotated for the presence of diagnostic cells. For a given slide, the frozen ThyVGG predicts for each patch the probability that it contains relevant cells. These probabilities are then used to add the top 1500 most relevant patches to our thyroid dataset. 

While cytology images are characterized by variations in preparation styles, staining intensities, scanner types, and other institutional differences, we do not apply any normalization techniques related to these differences. The pretraining method of our CytoFM inherently learns to ignore such subtle differences and noise, instead focusing on the cytology relevant information \cite{zhou_ibot_2022}.

%-------------------------------------------------------------------------

% \begin{figure*}[t]
%   \centering
%   %\fbox{\rule{0pt}{2in} \rule{0.9\linewidth}{0pt}}
%    %\includegraphics[width=0.8\linewidth]{cytofm.jpg}
%    \includegraphics[width=0.8\textwidth]{cytofm.jpg}

%    \caption{CytoFM. blah blah blah}
%    \label{fig:twocol}
% \end{figure*}

%-------------------------------------------------------------------------

%-------------------------------------------------------------------------
\begin{table*}[]
\centering
\begin{tabular}{c|cc|cc|cc}
              & \multicolumn{2}{c|}{FNAC2019}                 & \multicolumn{2}{c|}{MLBC}                     & \multicolumn{2}{c}{HiCervix}    \\ \hline
              & Acc.                  & AUC                   & Acc.                  & AUC                   & Acc.           & AUC            \\ \hline
iBOT-Imagenet & \textbf{0.946 ± 0.05} & \textbf{0.991 ± 0.01} & 0.879 ± 0.06          & 0.983 ± 0.01          & 0.803          & 0.956          \\
UNI           & 0.927 ± 0.06          & 0.983 ± 0.02          & 0.895 ± 0.06          & 0.986 ± 0.01          & 0.800          & 0.952          \\
CytoFM        & 0.908 ± 0.06          & 0.979 ± 0.02          & \textbf{0.930 ± 0.05} & \textbf{0.993 ± 0.01} & \textbf{0.844} & \textbf{0.968} \\ \hline
\end{tabular}
\caption{\textbf{Downstream performance} results of iBOT-Imagenet, UNI, and CytoFM on two cell type and one malignancy classification tasks.  
The best performance is in bold. HiCervix is a completely unseen dataset in the training of CytoFM while FNAC2019 and MLBC have a portion of their dataset seen and a portion completely unseen during CytoFM training.
}
\label{table1}
\end{table*}

%-------------------------------------------------------------------------

\subsection{ViT pre-training }
\label{subsec:preprocess}

For pretraining CytoFM, we use iBOT \cite{zhou_ibot_2022}, a state-of-the-art self supervised learning method for training ViTs using a teacher-student framework with self-distillation and masked image modeling (MIM) using an online tokenizer. iBOT uses self-distillation to learn a semantically-meaningful online tokenizer through minimizing the cross-entropy loss between the predictive categorical distributions from the teacher and student network. Given an image from the dataset, $x \thicksim \mathbb{D}$ and two randomly augmented views of the image $u$ and $v$, the cross-entropy loss is computed across views ($u$ compared to $v$) of the predicted classification tokens from the student, $u^{[CLS]}_s$ and $v^{[CLS]}_s$, and the teacher, $u^{[CLS]}_t$ and $v^{[CLS]}_t$. Self-distillation encourages the model to focus on the important information from an image while ignoring noise, such as variations in stain intensity. High-level context and features are learned by iBOT through MIM using self-distillation, where the model aims to predict masked areas of an image using the remaining contextual information. Specifically, the student network is given an augmented image with blockwise masking while the teacher network is given the augmented image. The student produces projections of the patch tokens for the masked image and iBOT aims to match these embeddings to those produced by the teacher from the unmasked augmented image. Through inferring missing patches, spatial relationships must be learned resulting in the model recognizing both low level details like local structures and cellular morphology as well as high-level semantics like cytoplasm clusters. Following iBOT pretraining, a standalone ViT can be initialized using weights from the teacher ViT, which is a more stable and generalized version of the student, and then frozen. The resulting ViT will be able to produce image embeddings that encode both large-scale and fine-grained patterns.

DINOv2 \cite{oquab_dinov2_2024} expands on iBOT and has shown superior performance on ImageNet and in histopathology \cite{chen_towards_2024} with stronger linear probe performance. iBOT was developed for pretraining ViT-Base and ViT-Large models using smaller datasets than DINOv2 which was developed for ViT-Large and ViT-Giant pretraining \cite{chen_towards_2024}. Given the smaller size of the cytology dataset, we choose to use ViT-Base with iBOT pretraining and initialize the model with ImageNet-1K iBOT pretrained weights. 

%With larger datasets, DINOv2 pretraining would be feasible and may improve the representation learning.

% %-------------------------------------------------------------------------
% \begin{table*}[]
% \centering
% \begin{tabular}{c|cc|cc|cc}
%               & \multicolumn{2}{c|}{FNAC2019}                 & \multicolumn{2}{c|}{MLBC}                     & \multicolumn{2}{c}{HiCervix}    \\ \hline
%               & Acc.                  & AUC                   & Acc.                  & AUC                   & Acc.           & AUC            \\ \hline
% iBOT-ImageNet & \textbf{0.946 ± 0.05} & \textbf{0.991 ± 0.01} & 0.879 ± 0.06          & 0.983 ± 0.01          & 0.803          & 0.956          \\
% UNI           & 0.927 ± 0.06          & 0.983 ± 0.02          & 0.895 ± 0.06          & 0.986 ± 0.01          & 0.800          & 0.952          \\
% CytoFM        & 0.908 ± 0.06          & 0.979 ± 0.02          & \textbf{0.930 ± 0.05} & \textbf{0.993 ± 0.01} & \textbf{0.844} & \textbf{0.968} \\ \hline
% \end{tabular}
% \caption{\textbf{Downstream performance} results of iBOT-Imagenet, UNI, and CytoFM on two cell type and one malignancy classification tasks.  
% The best performance is in bold. HiCervix is a completely unseen dataset in the training of CytoFM while FNAC2019 and MLBC have a portion of their dataset seen and a portion completely unseen during CytoFM training.
% }
% \label{table1}
% \end{table*}

% %-------------------------------------------------------------------------

%-------------------------------------------------------------------------

%% file: sec/3_experiments.tex
\section{Experiments and results}
\label{sec:results}
%-------------------------------------------------------------------------
\subsection{Implementation details}
\label{subsec:results_implementations}
iBOT initialized with weights from ImageNet-1K pretraining, is used to pretrain a ViT-Base model with $ \thicksim$1.4 million cytology patches from seven datasets spanning three organs and seven institutions. CytoFM is the resulting trained ViT which is frozen and used to extract features for downstream classification tasks.

CytoFM performance is compared against iBOT-ImageNet, an iBOT-Base model initialized with weights from pre-trianing only on ImageNet-1K. We also compare performance against the histopathology specific foundation model, UNI. No additional pre-training or fine-tuning was performed for either of these models and weights were taken from the respective data releases.

%Following pretraining, the weights from the teacher ViT are used as the initialization weights for a standalone ViT-Base model, CytoFM. We extract features using the frozen CytoFM and perform downstream classification tasks.

We perform downstream classification on three datasets. Two of the datasets, FNAC2019 and MLBC, are split into two non-overlapping sets, one set used during pre-training and the other completely unseen during pre-training and held out for downstream analysis. Malignancy classification is done on a held out set of patches from 100 images in the FNAC2019 dataset where there is an even split of benign and malignant examples. Four class cell type classification is performed on a held out set of patches from 140 samples from the MLBC dataset where there is an even split of cell types. The held out portions of both datasets were split into train, validation, and test sets with a 60:20:20 ratio stratified on the class label. The HiCervix dataset is completely held out from pretraining of the feature extractor model. Cell type classification with four cell types is performed. The train, validation, and test set splits used are those included in the public release of the dataset. Image level labels provided in the public release of the datasets are used in the classification tasks.

Each image in the testing datasets is patched with the same methods as the datasets used for pretraining. Stain normalization is not applied to any of the downstream testing datasets. Using the frozen CytoFM, we extract features from each patch. ABMIL is then used to aggregate the extracted features from patches of each image into a bag for each image using learned attention weights. A linear layer is then used to produce a final predicted probability for that image for the specified classification task. The ABMIL model is fully trainable and metrics are computed on the test set using the model with the best performance on the validation set. Hyperparameters for the ABMIL model are kept the same for each downstream task and across all feature extractor model comparisons.

% %-------------------------------------------------------------------------

\begin{figure}[t]
  \centering
   \includegraphics[width=0.9\linewidth]{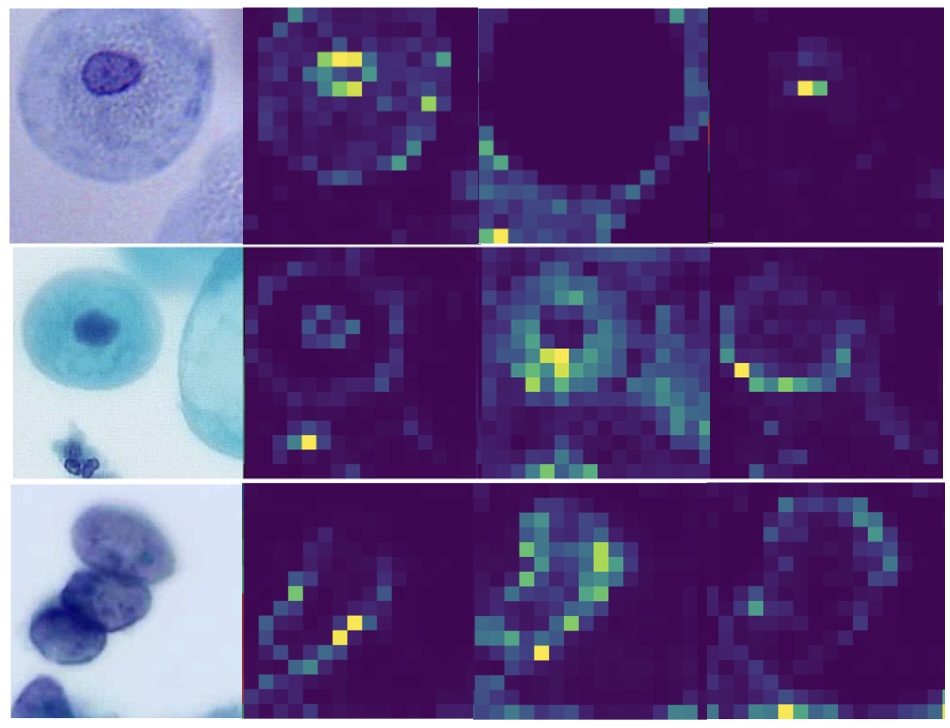}

   \caption{\textbf{CytoFM attentions}. Selected attention maps from the last layer of the model. The model attends to relevant cytological features such as the the nuclei, mitotic activity, nuclei boundaries, and morphology.}
   \label{fig:attns}
\end{figure}

%-------------------------------------------------------------------------

%-------------------------------------------------------------------------

\begin{figure*}[t]
  \centering
   \includegraphics[width=1\textwidth]{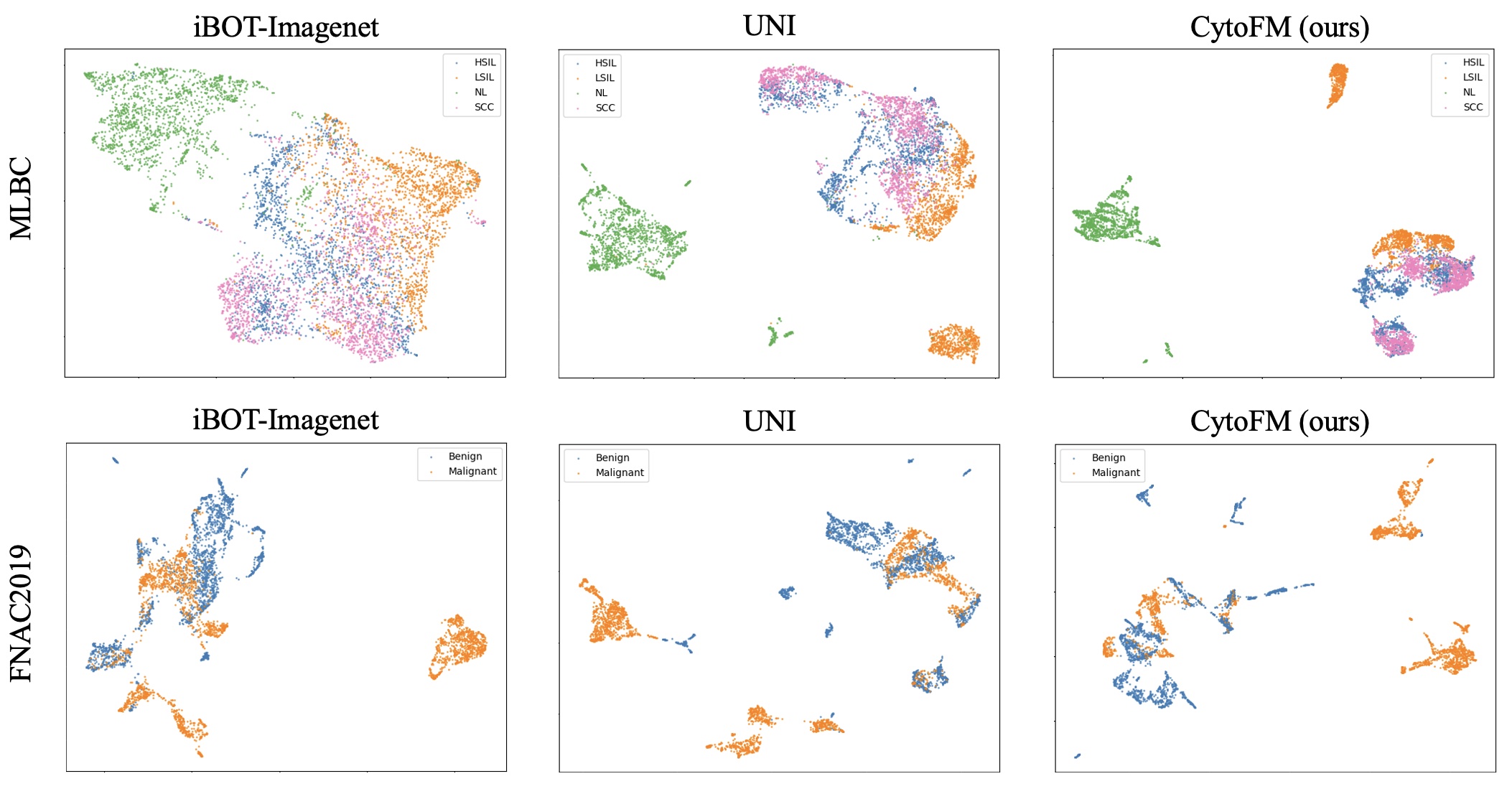}

   \caption{\textbf{UMAP of extracted features} for the MLBC (top row) and FNAC2019 (bottom row) datasets. Less dispersion in the features is seen in our model, CytoFM.}
   \label{fig:umap}
\end{figure*}

%-------------------------------------------------------------------------

% %-------------------------------------------------------------------------
% \begin{table*}[]
% \centering
% \begin{tabular}{c|cc|cc|cc}
%               & \multicolumn{2}{c|}{FNAC2019}                 & \multicolumn{2}{c|}{MLBC}                     & \multicolumn{2}{c}{HiCervix}    \\ \hline
%               & Acc.                  & AUC                   & Acc.                  & AUC                   & Acc.           & AUC            \\ \hline
% iBOT-ImageNet & \textbf{0.946 ± 0.05} & \textbf{0.991 ± 0.01} & 0.879 ± 0.06          & 0.983 ± 0.01          & 0.803          & 0.956          \\
% UNI           & 0.927 ± 0.06          & 0.983 ± 0.02          & 0.895 ± 0.06          & 0.986 ± 0.01          & 0.800          & 0.952          \\
% CytoFM        & 0.908 ± 0.06          & 0.979 ± 0.02          & \textbf{0.930 ± 0.05} & \textbf{0.993 ± 0.01} & \textbf{0.844} & \textbf{0.968} \\ \hline
% \end{tabular}
% \end{table*}

% % %-------------------------------------------------------------------------

% \begin{figure}[t]
%   \centering
%   %\fbox{\rule{0pt}{2in} \rule{0.9\linewidth}{0pt}}
%    \includegraphics[width=0.9\linewidth]{cytofm_attns.jpg}

%    \caption{CytoFM attentions. Selected attention maps from the last layer of the model. The model attends to relevant cytological features such as the the nuclei, mitotic activity, nuclei boundaries, and morphology.}
%    \label{fig:attns}
% \end{figure}

% %-------------------------------------------------------------------------

\subsection{Results}
\label{subsec:results}

For the binary classification task we report the accuracy at a threshold of 0.5 and the area under the receiver operating characteristic (AUROC) curve. In the case of multiclass classification, the top predicted probability is taken as the class prediction and the micro-averaged AUROC is reported. Due to the small size of the two partially held out datasets (FNAC2019 and MLBC), we train and evaluate the downstream model on 100 random train/test/validation splits and report the average over the metrics. For the large HiCervix dataset which is completely unseen during pretraining, we report only the metrics from the train/validation/test set specified by the dataset creators.

Classification performance on the held out portions of the FNAC2019 and MLBC datasets are shown in table \ref{table1}. CytoFM outperforms UNI and iBOT-Imagenet on MLBC, but not on FNAC2019. CytoFM performed the best in cell type classification of the MLBC dataset with an accuracy of 0.930 ± 0.05 (vs. UNI: $p<0.001$, vs iBOT-Imagenet: $p<0.001$) and an AUROC of 0.993 ± 0.01 (vs. UNI: $p<0.001$, vs iBOT-Imagenet: $p<0.001$). UNI and iBOT-Imagenet both had lower classification performance with accuracies of 0.895 ± 0.06 and 0.879 ± 0.06 and AUROC values of 0.986 ± 0.01 and 0.983 ± 0.01, respectively.  For malignancy classification, all three models achieve accuracies over 90 percent. iBOT-Imagenet had the highest accuracy, 0.946 ± 0.05 (vs. UNI: $p=0.026$, vs CytoFM: $p<0.001$), and AUROC, 0.991 ± 0.01 (vs. UNI: $p<0.004$, vs CytoFM: $p<0.001$). UNI and CytoFM both had lower performance with accuracies of 0.927 ± 0.06 and 0.908 ± 0.06 and AUROC values of 0.983 ± 0.02 and 0.979 ± 0.02, respectively. 

CytoFM had the highest performance on the classification task on the HiCervix dataset, which was completely unseen during model training. On this unseen dataset, CytoFM achieved an accuracy of 0.844 and an AUROC of 0.968. iBOT-Imagenet and UNI perform worse,  with accuracies of 0.803 and 0.800 and AUROCs of 0.956 and 0.952, respectively.

%-------------------------------------------------------------------------

\section{Discussion}
\label{sec:discussion}

Our model outperforms the non cytology foundation models on two of the three downstream tasks, and achieves high performance on the third task. The promising performance of CytoFM demonstrates that the model is learning relevant cytological details and the generalizability of our foundation model. 

To assess the ability of CytoFM to learn relevant features, we visualize the embeddings from each model for the two held out datasets using uniform manifold approximation projection (UMAP) (figure \ref{fig:umap}). Across the three models, we observe that CytoFM produces more defined clusters while the other models exhibit more dispersed feature distributions. The higher dispersion of features suggests that the non cytology-specific models may not have fully captured the cytology-specific features, particularly in the task of cell type classification. We also visualize attention maps (figure \ref{fig:attns}) from the final layer of CytoFM to evaluate which areas of cytology images the model attends to. The maps show that the model attends to characteristics such as cell morphology and areas of mitotic activity which are important in cytology classification. Therefore, cytology-specific foundation models are able to harness the complex information present in cytology images. In future work, we will explore the impact of attending to these cytology-specific features in downstream applications.

CytoFM outperforms iBOT-Imagenet and UNI on two out of the three downstream tasks. The lower performance on the FNAC2019 dataset compared to the other models may be a result of the disproportional amount of breast cytology in the pretraining of CytoFM. Due to limited dataset availability, only 2,544 images in the pretraining set represented breast cytology compared to close to a million cervix cytology images. Given the high performance on the cervix datasets, MLBC and HiCervix, the model likely became more cervix cytology focused. In future work, we will explore the use of resampling methods in pretraining that could improve the generalizability of the model to other domains \cite{jiao_usfm_2024} and ablations to evaluate dataset importance.
%\textcolor{red}{Future work will also include ablation studies to evaluate the importance of each dataset in robust learning representations.}

The performance of CytoFM on the HiCervix dataset is slightly lower compared to the current HiCervix state of the art which uses a HeirSwin and has an accuracy of 0.921. The HeirSwin was both trained and evaluated on the HiCervix dataset which limits the generalization ability of the model to external datasets. CytoFM, on the other hand, never sees the dataset when learning feature representations. In order to present a fair comparison with a focus on the features, we use a simple classification model and do not perform any calibration of the results or tune hyperparameters for the downstream tasks. These modifications could provide an increase in performance resulting in metrics more similar to the reported state of the art. Nevertheless, CytoFM still performs competitively and demonstrates the ability of our model to generate robust representations on unseen datasets from different institutions without the need for additional fine-tuning. The rich representations that CytoFM learns could be leveraged in multimodal applications, enabling the integration of cytology with other data modalities, such as omics and molecular testing to enhance diagnostic accuracy and clinical insights.

%\textcolor{red}{Include some limitations.}

% limited downstream tasks
% one comment of using more partial data, include in downstream? or maybe not mention, can think how might mention
% more ablations, like droppign datasets, including more qhatnot
% more emphasis i guess on the parameter tuning and fine tuning which could get better and more comparable results especially compared to the specific models like HeirSwin. our benefit which maybe emphasize more is that we dont need more data for training the model and sure with fine tuning there could be improvements to bring it to heirswin level which we will look at in future work. but we show it is doing well with absolutely no fine tuning.
% the 'cytology specific' foundation models 

%-------------------------------------------------------------------------

%% file: sec/4_conclusions.tex
\section{Conclusions}
\label{sec:conclusions}

CytoFM is the first cytology foundation model. The model is trained in a self-supervised manner using a multi-institutional and multi-organ dataset containing $ \thicksim$1.4 million patches. In two out of three downstream tasks, our model outperforms foundation models trained on natural images and histology. We highlight that our model learns robust and meaningful cytology-specific features which generalize to unseen datasets. Given the competitive performance of CytoFM, we will explore additional training techniques and more extensive ablations and evaluations in future work. CytoFM's performance demonstrates the promise that cytology foundation models can act as robust and generalizable feature extractors, crucial for the application of deep learning to cytology. 